\documentclass[10pt,twocolumn,letterpaper]{article}

\usepackage{multirow}
\usepackage{algorithm}
\usepackage{algorithmic}
\usepackage[accsupp]{axessibility} 
\usepackage[pagenumbers]{cvpr} 

\newtheorem{theorem}{Theorem}[section]

\definecolor{cvprblue}{rgb}{0.21,0.49,0.74}
\usepackage[pagebackref,breaklinks,colorlinks,allcolors=cvprblue]{hyperref}
\usepackage{xcolor}
\usepackage{colortbl}
\usepackage{graphicx}
\def\paperID{} 
\def\confName{CVPR}
\def\confYear{2026}

\title{CRAFT: Aligning Diffusion Models with Fine-Tuning Is Easier Than You Think}

\author{
\textbf{Zening Sun}$^{1}$\thanks{Equal contribution.} \quad
\textbf{Zhengpeng Xie}$^{1}$\footnotemark[1] \quad
\textbf{Lichen Bai}$^{1}$ \quad
\textbf{Shitong Shao}$^{1}$ \quad
\textbf{Shuo Yang}$^{2}$ \quad
\textbf{Zeke Xie}$^{1}$\thanks{Corresponding author.} \\
$^{1}$ HKUST (GZ) \quad
$^{2}$ HIT (SZ)
}

\begin{document}
\maketitle
\begin{abstract}
Aligning Diffusion models has achieved remarkable breakthroughs in generating high-quality, human preference-aligned images. Existing techniques, such as supervised fine-tuning (SFT) and DPO-style preference optimization, have become principled tools for fine-tuning diffusion models. 
However, SFT relies on high-quality images that are costly to obtain, while DPO-style methods depend on large-scale preference datasets, which are often inconsistent in quality. 
Beyond data dependency, these methods are further constrained by computational inefficiency. To address these two challenges, we propose Composite Reward Assisted Fine-Tuning (CRAFT), a lightweight yet powerful fine-tuning paradigm that requires significantly reduced training data while maintaining computational efficiency. It first leverages a Composite Reward Filtering (CRF) technique to construct a high-quality and consistent training dataset and then perform an enhanced variant of SFT. We also theoretically prove that CRAFT actually optimizes the lower bound of group-based reinforcement learning, establishing a principled connection between SFT with selected data and reinforcement learning. Our extensive empirical results demonstrate that CRAFT with only 100 samples can easily outperform recent SOTA preference optimization methods with thousands of preference-paired samples. Moreover, CRAFT can even achieve 11-220$\times$ faster convergences than the baseline preference optimization methods, highlighting its extremely high efficiency.
\end{abstract}


\section{Introduction}

\begin{figure}[t]
    \centering
    \includegraphics[width=\columnwidth]{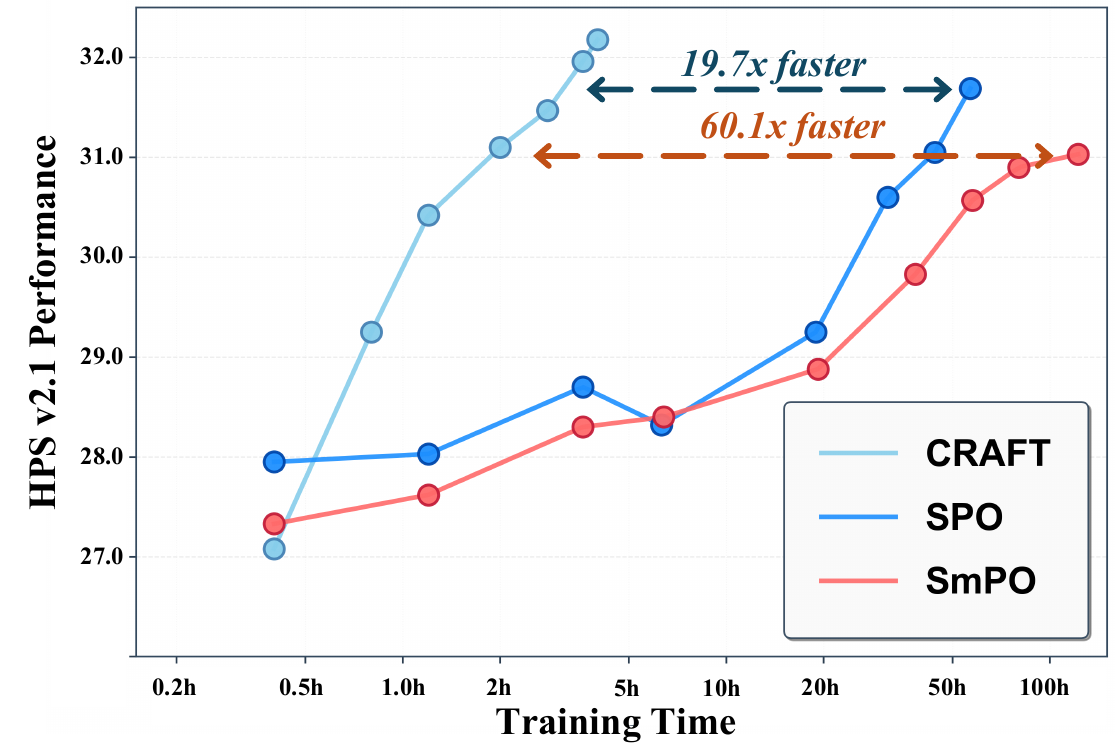}
    \caption{\textbf{Training Efficiency Comparison.} Compared to SPO and SmPO, CRAFT reaches the same HPSv2.1 performance with 19.7× and 60.1× faster training time respectively, and further achieves superior final performance, demonstrating a dual advantage in both training speed and final generation quality.}
    \label{fig:efficiency_comparison}
\end{figure}

\begin{figure*}[!t]
	\centering
	\includegraphics[width=\textwidth]{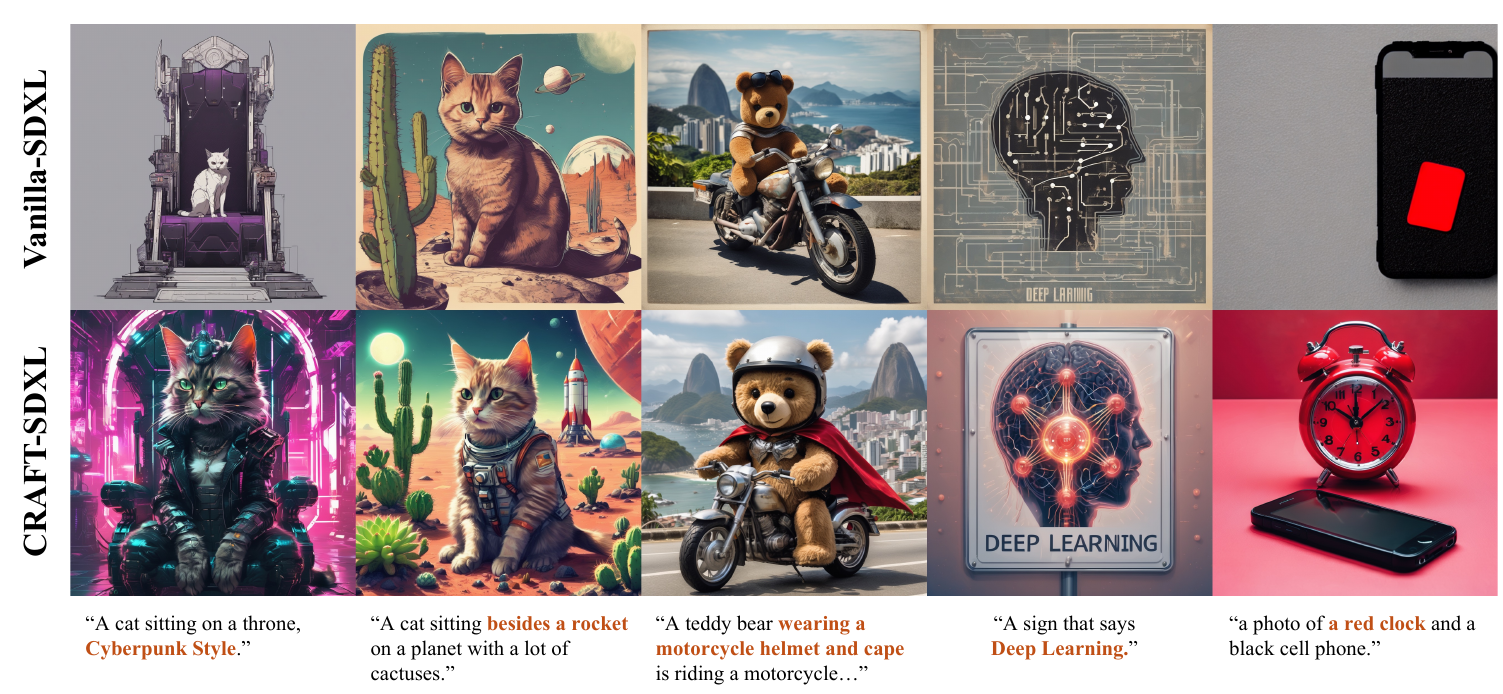}
	\caption{\textbf{Qualitative Improvements over Vanilla SDXL.} We present CRAFT, an efficient and effective fine-tuning method designed to enhance diffusion models' alignment with human preference. The top row displays images generated by the base Vanilla-SDXL model, while the bottom row shows results generated by our CRAFT-finetuned model (CRAFT-SDXL). The comparison clearly shows that CRAFT-SDXL excels at complex instruction following and compositional reasoning, demonstrating significant superiority in adhering to diverse stylistic concepts (e.g., ``Cyberpunk Style”), accurately generating specified objects and their attributes (e.g., the position of the cat ``beside a rocket" and the teddy bear ``wearing a helmet and cape"), and precisely rendering on-image text (e.g., ``Deep Learning”).}
    \label{fig:intro_qualitative}
\end{figure*}

Recent advances in diffusion models (DMs)~\cite{ho2020denoising, song2020score} have significantly propelled text-to-image (T2I) generation, achieving substantial improvements in visual fidelity and text-image alignment~\cite{rombach2022high, podell2023sdxlimprovinglatentdiffusion, labs2025flux}. Despite these breakthroughs, diffusion models are typically pretrained on noisy large-scale, web-sourced datasets~\cite{schuhmann2022laion, saharia2022photorealistic, ramesh2021zero}. Therefore, their outputs often fail to align with human preferences for aesthetics and instruction-following, highlighting a need for effective alignment. Inspired by the success of post-training techniques in large language models (LLMs)~\cite{ouyang2022training, wei2021finetuned, rafailov2023direct}, numerous recent studies have adapted these methods to fine-tune diffusion models for human preference alignment.

Existing post-training approaches present a trade-off between data requirements and computational cost. Supervised Fine-Tuning (SFT), while computationally lean, hinges on access to exceptionally high-quality datasets that are scarce and expensive to curate. Inspired by the success of Direct Preference Optimization (DPO) in LLMs~\cite{rafailov2023direct}, which learns directly from large-scale preference pairs, Diffusion-DPO~\cite{wallace2024diffusion} adapted this paradigm to fine-tune diffusion models. This offline approach achieves strong alignment and inspired many DPO-style methods. A prominent example is SmPO~\cite{lu2025smoothed}, which builds upon the DPO framework by incorporating smoothed reward signal processing. While effective, the fundamental limitation of these offline DPO-style approaches is their reliance on large-scale data. Alternatively, many online methods, exemplified by SPO~\cite{liang2025aesthetic}, generate preference pairs dynamically at each denoising step. Although this avoids reliance on large-scale datasets, it incurs immense computational overhead from repeated online sampling and evaluation. Consequently, aligning diffusion models remains challenging due to the scarcity of high-quality data and the high computational cost of existing methods. Therefore, a fine-tuning approach that is both data-efficient and computationally lightweight is important for the future of diffusion model alignment.

To address these two challenges, we propose \textbf{Composite Reward Assisted Fine-Tuning (CRAFT)}, a simple yet effective framework for aligning diffusion models efficiently with as few as 100 samples. CRAFT introduces a self-curated and computationally efficient training paradigm that fine-tunes models through an enhanced SFT guided by a composite reward function. Specifically, we leverage our Composite Reward Filtering strategy, which enables the automatic selection of representative, distribution-consistent training samples without reliance on external high-quality datasets or data distilled from stronger models. Moreover, we theoretically prove that CRAFT can be interpreted as a lower bound of group-based reinforcement learning, providing a principled bridge between supervised fine-tuning and reinforcement learning from human feedback.

We use CRAFT to fine-tune Stable Diffusion v1.5 and Stable Diffusion XL using only \textbf{100} samples. As illustrated in Fig.~\ref{fig:efficiency_comparison}, CRAFT reaches the same HPSv2.1 performance 19.7× faster than SPO and 60.1× faster than SmPO and achieves the best HPSv2.1 performance. Moreover, as shown in Fig.~\ref{fig:intro_qualitative} and Fig.~\ref{fig:qualitative_results}, CRAFT substantially enhances the generative quality and aesthetic fidelity. Notably, our composite reward function does not include ImageReward and MPS, yet CRAFT attains the highest scores on SDXL and SD1.5 across all tests (see Table~\ref{sdxl_main} and Table~\ref{SD1.5_main}). Our main contributions are summarized as follows:\\
\textbf{(i)} We propose CRAFT, a simple yet effective alignment framework that eliminates the dependence on large-scale preference datasets while reducing computational cost, achieving strong preference alignment with 100 samples.\\
\textbf{(ii)} We theoretically prove that our CRAFT can be interpreted as a lower-bound optimization of group-based reinforcement learning, establishing a principled connection between supervised fine-tuning and reinforcement learning.\\
\textbf{(iii)} Extensive experiments on SD1.5 and SDXL demonstrate that our CRAFT achieves up to 11–220× faster training while maintaining superior preference alignment, confirming the efficiency and effectiveness of our approach.

\begin{figure*}[!t]
	\centering
	\includegraphics[width=\textwidth]{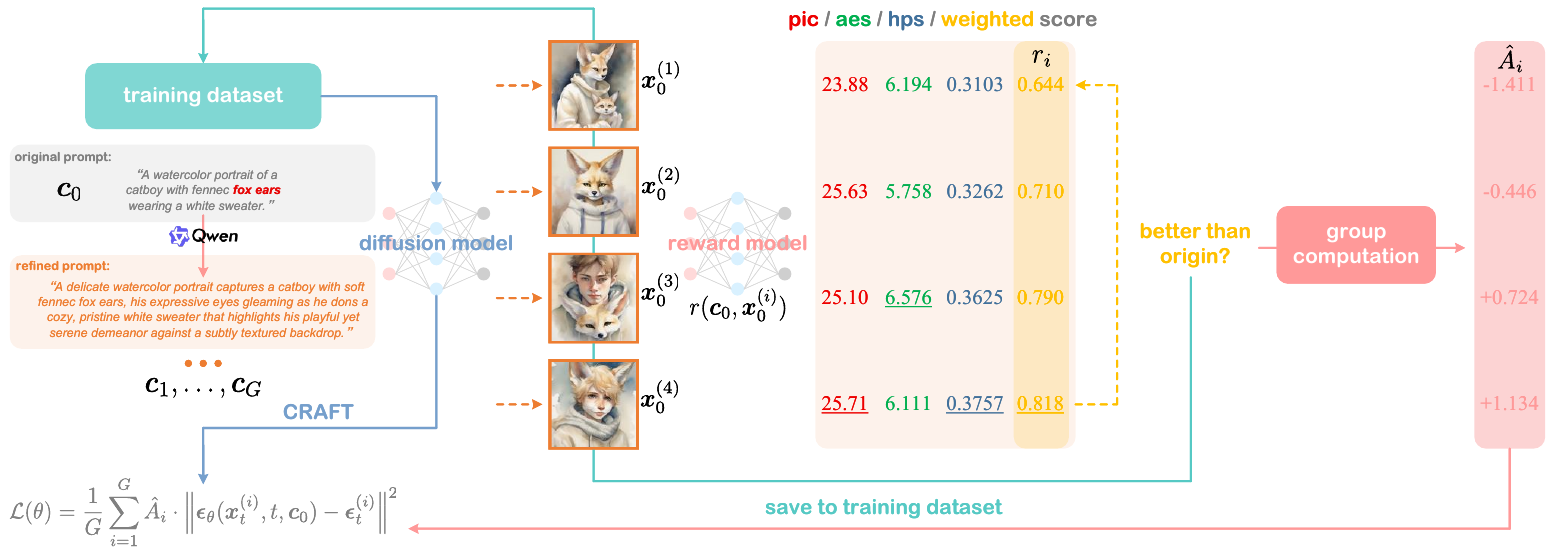}
	\caption{\textbf{The overall pipeline of our method} mainly consists of three stages: (i) data construction, (ii) composite reward filtering, and (iii) weighted SFT fine-tuning.}
\end{figure*}

\section{Related Work}

\paragraph{Text-to-Image Diffusion Models.}
Diffusion models (DMs)~\cite{yang2023diffusion, song2020score, ho2020denoising, song2020improved, liu2026alignment} have emerged as the dominant paradigm for text-to-image (T2I) generation, surpassing earlier models such as GANs~\cite{reed2016generative, karras2019style} in visual fidelity. Building upon the foundational advances in denoising diffusion probabilistic models, DALL-E~\cite{ramesh2021zero}, Imagen~\cite{saharia2022photorealistic}, and Stable Diffusion~\cite{rombach2022high, podell2023sdxlimprovinglatentdiffusion} introduced powerful text-conditioning mechanisms and latent-space diffusion, enabling high-resolution and prompt-controllable image synthesis. However, although producing visually compelling results, current T2I diffusion models often fail to align closely with human preferences~\cite{wallace2024diffusion, li2024aligning, yang2023diffusion}. This misalignment primarily arises from their training stage, which rely on large-scale but noisy datasets (e.g., LAION-5B~\cite{schuhmann2022laion}) and simple reconstruction losses~\cite{ho2020denoising, rombach2022high}. In this work, we introduce a simple yet effective method CRAFT to make fine-tuning diffusion models easier.
\paragraph{Aligning Diffusion Models.}
Inspired by post-training techniques in large language models (LLMs) such as SFT and RLHF, diffusion model alignment has recently gained increasing attention. Supervised Fine-Tuning (SFT) directly fine-tunes diffusion models on curated text–image pairs to improve prompt adherence and visual fidelity, yet its reliance on high-quality data limits practicality. RL-based alignment adapts RLHF methods by framing denoising as a policy optimization process. Approaches such as DDPO~\cite{black2023training}, DPOK~\cite{fan2023dpok}, and GRPO-style extensions~\cite{xue2025dancegrpo, liu2025flow, shao2024deepseekmath} introduce log-probability based supervision for stability, but remain computationally expensive and often unstable under sparse or noisy rewards.
Recently, preference optimization has emerged as a more practical alternative for aligning diffusion models with human judgments. Direct Preference Optimization (DPO)~\cite{rafailov2023direct} replaces explicit reward modeling with pairwise preference supervision, achieving alignment comparable to RLHF in LLMs. This concept was later extended to diffusion models via Diffusion-DPO~\cite{wallace2024diffusion}, which directly optimizes generation quality using preference pairs. Building on this paradigm, subsequent variants, such as DSPO~\cite{zhu2025dspo}, SPO~\cite{liang2025aesthetic}, and SmPO~\cite{lu2025smoothed} explore score coupling, step-aware rewards, and reward smoothing for improved stability and fidelity. 

\section{Methodology}
In this section, we first build some theoretical insights in Section \ref{Formulation and Theoretical Insights}, introduce \textit{Composite Reward Filtering} (CRF) technique in Section \ref{Composite Reward Filtering}, and in Section \ref{Composite Reward Assisted Fine-Tuning} we propose our practical fine-tuning algorithm called \textit{Composite Reward Assisted Fine-Tuning} (CRAFT).

\subsection{Formulation and Theoretical Insights}\label{Formulation and Theoretical Insights}
We begin by introducing some basic formulation under reinforcement learning settings. Suppose we have a diffusion model $p_\theta(\boldsymbol{x}_0|\boldsymbol{c})$, where $\boldsymbol{c}\in\mathcal{D}_{\boldsymbol{c}}$ is the prompt, and $\boldsymbol{x}_0$ is the clean image. Given a reward model $r(\boldsymbol{c},\boldsymbol{x}_0)$, our goal is to optimize the following objective:
\begin{equation}
    \begin{split}
    J(\theta)&=\mathbb{E}_{\boldsymbol{c}\sim\mathcal{D}_{\boldsymbol{c}},\boldsymbol{x}_0\sim p_\theta(\cdot|\boldsymbol{c})}\left[r(\boldsymbol{c},\boldsymbol{x}_0)\right]\\
        &=\mathbb{E}_{\boldsymbol{c}\sim\mathcal{D}_{\boldsymbol{c}},\boldsymbol{x}_0\sim p_{\theta_{\mathrm{old}}}(\cdot|\boldsymbol{c})}\left[\frac{p_{\theta}(\boldsymbol{x}_0|\boldsymbol{c})}{p_{\theta_{\mathrm{old}}}(\boldsymbol{x}_0|\boldsymbol{c})}\cdot r(\boldsymbol{c},\boldsymbol{x}_0)\right],
    \end{split}
\end{equation}
where we use \textit{importance sampling} to decouple the behavior policy $p_{\theta_{\mathrm{old}}}$ and the target policy $p_{\theta}$.
While traditional RLHF formulations employ KL regularization to prevent reward over-optimization, we omit this component since prior works indicate only minimal performance differences \citep{xue2025dancegrpo}. Then, motivated by the core idea of group-based fine-tuning technique \citep{shao2024deepseekmath}, we can further approximate this objective using the \textit{Monte Carlo} method as follows:
\begin{equation}
    \hat{J}(\theta)=\mathbb{E}_{\boldsymbol{c}\sim\mathcal{D}_{\boldsymbol{c}},\{\boldsymbol{x}_0^{(i)}\}_{i=1}^G\sim p_{\theta_{\mathrm{old}}}(\cdot|\boldsymbol{c})}\left[\frac{1}{G}\sum_{i=1}^{G}\frac{p_{\theta}(\boldsymbol{x}_0^{(i)}|\boldsymbol{c})}{p_{\theta_{\mathrm{old}}}(\boldsymbol{x}_0^{(i)}|\boldsymbol{c})}\cdot\hat{A}_i\right],
\end{equation}
where $\hat{A}_i=(r(\boldsymbol{c},\boldsymbol{x}_0^{(i)})-\mathrm{mean})/(\mathrm{std}+\epsilon)$, $\mathrm{mean}$ and $\mathrm{std}$ represent the mean and standard deviation of the group rewards $\{r(\boldsymbol{c},\boldsymbol{x}_0^{(i)})\}_{i=1}^{G}$, and $\epsilon>0$ is a small constant for numerical stability. 

We are now ready to present the main theoretical result of this paper:
\begin{theorem}\label{lower bound}
    Given a Text-to-Image (T2I) diffusion model $p_\theta(\boldsymbol{x}_0|\boldsymbol{c})$, for any gradient direction $g$ with a small learning rate\footnote{Such assumption is reasonable, since most fine-tuning algorithms set the learning rate between 1e-5 and 1e-6.} $\eta\rightarrow0$ ($\theta=\theta_{\mathrm{old}}+\eta g$), the following bound holds:
    \begin{equation*}
    \begin{split}
        &\hat{J}(\theta)\geq C-\\
        &\mathop{\mathbb{E}}_{\substack{\boldsymbol{c}\sim\mathcal{D}_{\boldsymbol{c}},\\\{\boldsymbol{x}_0^{(i)}\}_{i=1}^G\sim p_{\theta_{\mathrm{old}}}(\cdot|\boldsymbol{c}),\\t\sim\mathrm{Uniform}(\{1,\dots,T\})\\\boldsymbol{\epsilon}_t^{(i)}\sim\mathcal{N}(0,I)}}\left[\frac{1}{G}\sum_{i=1}^{G}\hat{A}_iw(t)\cdot\left\Vert\boldsymbol{\epsilon}_\theta(\boldsymbol{x}_t^{(i)},t,\boldsymbol{c})-\boldsymbol{\epsilon}_t^{(i)}\right\Vert^2\right],\\
    \end{split}
    \end{equation*}
    where $\boldsymbol{\epsilon}_{\theta}$ is the noise predictive model of $p_{\theta}$, and $C$ is a constant that do not depends on $\theta$.
\end{theorem}
The detailed proof of Theorem \ref{lower bound} is shown in Appendix. Theorem \ref{lower bound} establishes a principled connection between \textit{group-based RL} and \textit{SFT}. Intuitively, given a batch of data for fine-tuning diffusion models (which can be either model-generated or existing data, corresponding to the \textit{online} and \textit{offline} settings, respectively), the data quality (i.e., the normalized advantage $\hat{A}_i$) determines the magnitude of the gradient, encouraging the model to generate high-quality images while avoiding low-quality ones.

\begin{algorithm}[!t]
\caption{\textit{Composite Reward Assisted Fine-Tuning}} \label{CRAFT algorithm}
\begin{algorithmic}[1]
    \REQUIRE Pre-trained Text-to-Image (T2I) diffusion model $p_\theta(\boldsymbol{x}_0|\boldsymbol{c})$, original prompt set $\mathcal{P}$ and refined prompt set $\mathcal{P}_{\mathrm{refined}}$, refined image dataset $\mathcal{I}=\cup_{i,j}\{\boldsymbol{x}_0^{(i,j)}\}$, reward functions $r_h,r_p,r_a$, reward filtering rules $\tilde{\mathcal{I}}\in\{\mathcal{I}_{h},\mathcal{I}_{p},\mathcal{I}_{a},\mathcal{I}_{ha},\mathcal{I}_{pa},\mathcal{I}_{hpa}\}$, batch $b\geq1$
    \REPEAT
        \STATE Randomly sample a small batch index $i=k,k+1,\dots,k+b-1$ from $\mathcal{I}$
        \STATE Compute group-based advantage:
        \begin{equation*}
            \hat{A}^{(i,j)}=\frac{r_{\mathrm{total}}^{(i,j)}-\mathrm{mean}\{r_{\mathrm{total}}^{(i,1)},\dots,r_{\mathrm{total}}^{(i,N)}\}}{\mathrm{std}\{r_{\mathrm{total}}^{(i,1)},\dots,r_{\mathrm{total}}^{(i,N)}\}+\epsilon}
        \end{equation*}
        \STATE Compute weighted-SFT loss:
        \begin{equation*}
        \begin{split}
            \mathcal{L}(\theta):=&\frac{1}{bN}\sum_{i=k}^{k+b-1}\sum_{j=1}^{N}\\
            \hat{A}^{(i,j)}&\cdot\left\Vert\boldsymbol{\epsilon}_\theta(\boldsymbol{x}_t^{(i,j)},t,\boldsymbol{c}_i^{(0)})-\boldsymbol{\epsilon}_t^{(i,j)}\right\Vert^2\cdot\mathbb{I}(\boldsymbol{x}_0^{(i,j)}\in\tilde{\mathcal{I}})\\
        \end{split}
        \end{equation*}
        \STATE Optimize $\mathcal{L}$ using gradient decent algorithm
    \UNTIL{converged}
\end{algorithmic}
\end{algorithm}

\subsection{Composite Reward Filtering}\label{Composite Reward Filtering}
\paragraph{Data construction.} We now provide our data filtering strategy in detail. We randomly sample $n=10000$ original prompts from the HPD dataset, denoted as $\mathcal{P}=\{\boldsymbol{c}_1^{(0)},\boldsymbol{c}_2^{(0)},\dots,\boldsymbol{c}_n^{(0)}\}$, and then we use Qwen-Plus to refine each prompt $N$ times, resulting in a set of refined prompts $\mathcal{P}_{\mathrm{refined}}=\{\boldsymbol{c}_1^{(1)},\dots,\boldsymbol{c}_1^{(N)},\dots,\boldsymbol{c}_n^{(1)},\dots,\boldsymbol{c}_n^{(N)}\}$. Next, we are ready to build our image dataset. We use the base model $p_\theta(\boldsymbol{x}_0|\boldsymbol{c})$ to generate images corresponding to all $n(N+1)$ prompts in set $\mathcal{P}\cup\mathcal{P}_{\mathrm{refined}}$, obtaining the original image dataset $\mathcal{I}=\cup_{\boldsymbol{c}\in\mathcal{P}\cup\mathcal{P}_{\mathrm{refined}}}\{\boldsymbol{x}_0\sim p_\theta(\cdot|\boldsymbol{c})\}$. We further use the notation $\boldsymbol{x}_0^{(i,j)}\sim p_\theta(\cdot|\boldsymbol{c}_i^{(j)})$ to distinguish different images, so $\mathcal{I}=\cup_{i,j}\{\boldsymbol{x}_0^{(i,j)}\}$, where $i=1,\dots,n;j=0,\dots,N$.

\paragraph{Data filtering.} To introduce our composite-reward filtering (CRF) strategy, we employ three reward models, they are HPS v2.1~\cite{wu2023human}, PickScore~\cite{kirstain2023pick} and AES metric~\cite{schuhmann2022laion}. We denote these models as $r_h(\boldsymbol{c},\boldsymbol{x}_0)$, $r_p(\boldsymbol{c},\boldsymbol{x}_0)$, and $r_a(\boldsymbol{c},\boldsymbol{x}_0)$, respectively. Note that for all images $\boldsymbol{x}_0\in\mathcal{I}$, we compute rewards using their \textit{original} prompts, and refined prompts $\mathcal{P}_{\mathrm{refined}}$ is solely to enhance the diversity of generated images. Formally,
\begin{equation}
    r_\xi^{(i,j)}=r_\xi(\boldsymbol{c}_i^{(0)},\boldsymbol{x}_0^{(i,j)}),
\end{equation}
where $r_\xi^{(i,j)}$ simply denotes the reward for each image $\boldsymbol{x}_0^{(i,j)}$ given the reward model $r_\xi$, $\xi$ can be chosen as $h$, $p$, or $a$. With these preparations, we introduce the following three different levels of filtering strategies:
\begin{equation}
    \begin{split}
        &\mathcal{I}_{\xi}=\left\{\boldsymbol{x}_0^{(i,j)}\in\mathcal{I}\middle\vert\sum_{j=1}^{N}\mathbb{I}\left(r_\xi^{(i,j)}>r_\xi^{(i,0)}\right)\neq0\right\};\\
        &\mathcal{I}_{ha}=\left\{\boldsymbol{x}_0^{(i,j)}\in\mathcal{I}\middle\vert\sum_{j=1}^{N}\prod_{\xi\in\{h,a\}}\mathbb{I}\left(r_\xi^{(i,j)}>r_\xi^{(i,0)}\right)\neq0\right\}; \\
        &\mathcal{I}_{pa}=\left\{\boldsymbol{x}_0^{(i,j)}\in\mathcal{I}\middle\vert\sum_{j=1}^{N}\prod_{\xi\in\{p,a\}}\mathbb{I}\left(r_\xi^{(i,j)}>r_\xi^{(i,0)}\right)\neq0\right\}; \\
        &\mathcal{I}_{hpa}=\left\{\boldsymbol{x}_0^{(i,j)}\in\mathcal{I}\middle\vert\sum_{j=1}^{N}\prod_{\xi\in\{h,p,a\}}\mathbb{I}\left(r_\xi^{(i,j)}>r_\xi^{(i,0)}\right)\neq0\right\},\\
    \end{split}
\end{equation}

where $\mathbb{I}$ represents the indicator function. In other words, $\mathcal{I}_\xi$ represents the filtering rule for individual rewards: as long as the refined prompt produces an image with a higher reward than the original one, this batch of generated samples is retained. In contrast, $\mathcal{I}_{ha}$ and $\mathcal{I}_{pa}$ suppose that two types of rewards are simultaneously higher than the originals, while $\mathcal{I}_{hpa}$ requires all three rewards to be higher, making it the strictest rule.

\subsection{Composite Reward Assisted Fine-Tuning}\label{Composite Reward Assisted Fine-Tuning}
Given the theoretical insights from Section \ref{Formulation and Theoretical Insights} and the reward filtering techniques described in Section \ref{Composite Reward Filtering}, we now introduce our fine-tuning algorithm CRAFT, as shown in Algorithm \ref{CRAFT algorithm}. 

We use $r_{\mathrm{total}}^{(i,j)}$ to represent the weighted reward when more than one reward model is used for data filtering, i.e., the filtered datasets $\mathcal{I}_{h}$, $\mathcal{I}_{p}$,  $\mathcal{I}_{ha}$, and $\mathcal{I}_{hpa}$. We use all the original data and compute the group-based advantage $\hat{A}^{(i,j)}$ for each image $\boldsymbol{x}_0^{(i,j)}$. Then, we calculate the weighted-SFT loss in a small batch. We employ the indicator function $\mathbb{I}(\boldsymbol{x}_0^{(i,j)}\in\tilde{\mathcal{I}})$ to represent that only the filtered data are used for gradient computation. While in Theorem \ref{lower bound} the loss coefficient includes the term $w(t)$, in practice previous studies found that removing this term makes the implementation simpler and yields better results \citep{ho2020denoising}. Therefore, we also omit $w(t)$ and use the pure advantage value to represent the gradient magnitude.

\section{Experiment}
\label{sec:exp}


\begin{figure*}[t]
	\centering
	\includegraphics[width=\textwidth]{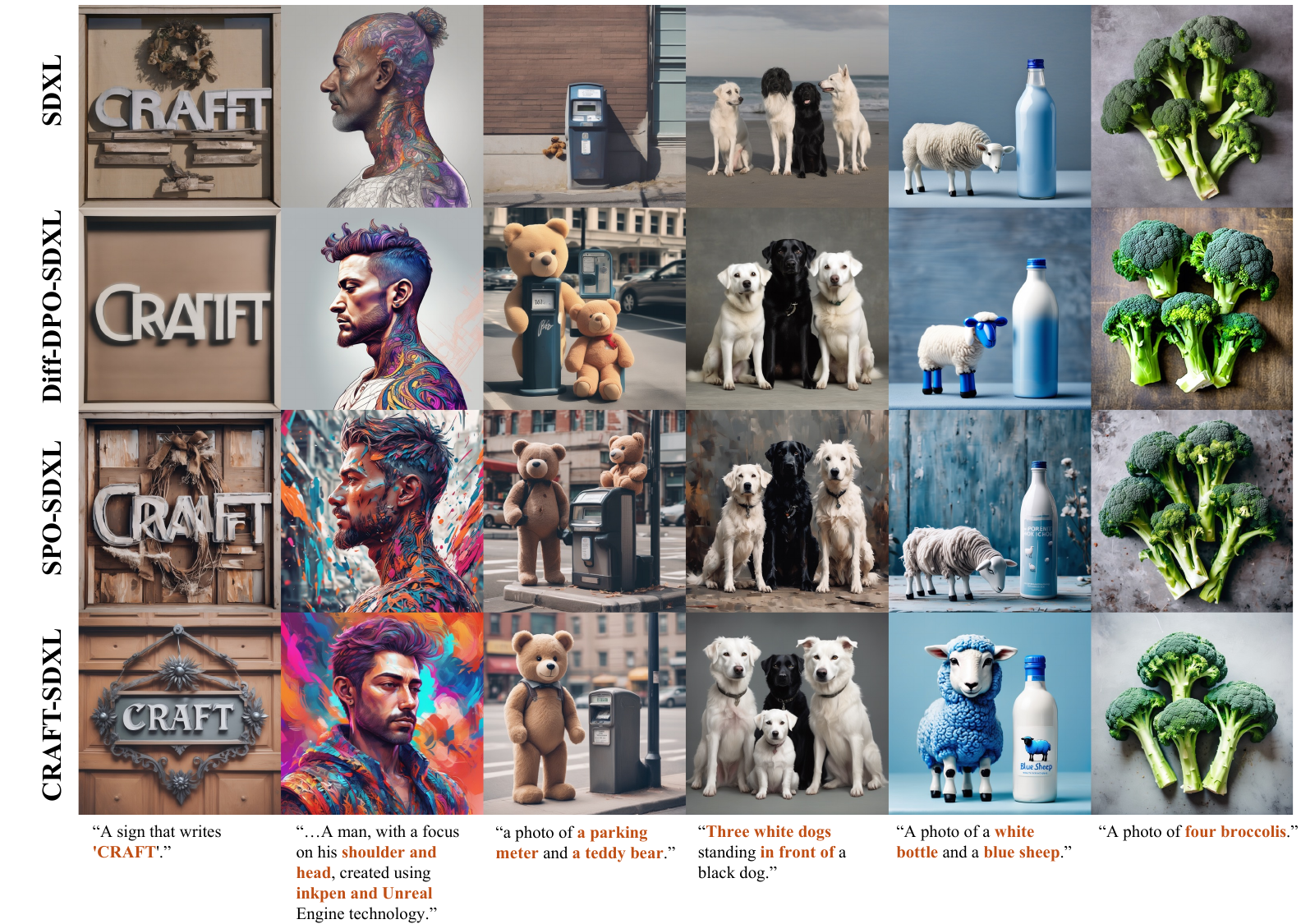}
    \caption{\textbf{Qualitative comparison.} We provide a qualitative comparison of CRAFT-Diffusion and other different preference optimization methods (Vanilla, Diff-DPO, SPO) for SDXL. CRAFT-SDXL generates images of superior quality and prompt fidelity, showcasing significant improvements in detail, composition, and text rendering.}
    \label{fig:qualitative_results}
\end{figure*}

\subsection{Experiment Setting}


\begin{table*}[t]
\centering
\caption{\textbf{Computational Cost Comparison.} We report the total NVIDIA H100 GPU hours for the fine-tuning stage. Our CRAFT framework demonstrates a dramatic reduction in computational cost, achieving significant speedups over all baselines.}
\label{tab:computation_cost_large}
\renewcommand{\arraystretch}{1} 
\setlength{\tabcolsep}{7pt}
\begin{tabular}{@{}lccc|lccc@{}}
\toprule
\multicolumn{4}{c|}{\textbf{SDXL}} & \multicolumn{4}{c}{\textbf{SD1.5}} \\
\cmidrule(r){1-4} \cmidrule(l){5-8}
\textbf{Method} & \textbf{Data} & \textbf{GPU Hours} & \textbf{Cost Ratio} & \textbf{Method} & \textbf{Data} & \textbf{GPU Hours} & \textbf{Cost Ratio} \\
\midrule
Diff-DPO & 851,293 & $\sim$638.1 & \textbf{159.5$\times$} & Diff-DPO & 851,293 & $\sim$80.6 & \textbf{42.4$\times$} \\
MaPO     & 928,068 & $\sim$545.6 & \textbf{136.4$\times$} & Diff-KTO & 851,293 & $\sim$415.6 & \textbf{218.7$\times$} \\
SmPO     & 851,293 & $\sim$120.1 & \textbf{30.0$\times$} & SmPO     & 851,293 & $\sim$27.6 & \textbf{14.5$\times$} \\
SPO      & 4,000   & $\sim$63.0  & \textbf{15.8$\times$} & SPO      & 4,000   & $\sim$22.0 & \textbf{11.6$\times$} \\
\textbf{CRAFT (Ours)} & \textbf{100} & $\sim$\textbf{4.0} & \textbf{1.0$\times$} & \textbf{CRAFT (Ours)} & \textbf{100} & $\sim$\textbf{1.9} & \textbf{1.0$\times$} \\
\bottomrule
\end{tabular}
\end{table*}

\begin{table*}[t]
\centering
\caption{\textbf{Quantitative Results} of our \textbf{CRAFT} and other baselines for SDXL. We report the average reward scores of images generated on HPDv2, Parti-Prompt, and Pick-a-Pic datasets. The highest value is shown in \textbf{bold}, the second highest is \underline{underlined}, and models marked with $^*$ are reproduced strictly following the official code and datasets.}
\label{sdxl_main}
\small
\setlength{\tabcolsep}{2pt}
    \begin{tabular}{c|cccc|cccc|cccc}
    \toprule
    \multirow{2}{*}{\textbf{Model}} & \multicolumn{4}{c|}{\textbf{HPDv2}} & \multicolumn{4}{c|}{\textbf{Parti-Prompt}} & \multicolumn{4}{c}{\textbf{Pick-a-Pic}} \\
    \cmidrule(lr){2-5} \cmidrule(lr){6-9} \cmidrule(lr){10-13}
    & HPSv2.1 $\uparrow$ & AES $\uparrow$ & ImgReward $\uparrow$ & MPS $\uparrow$ & HPSv2.1 $\uparrow$ & AES $\uparrow$ & ImgReward $\uparrow$ & MPS $\uparrow$ & HPSv2.1 $\uparrow$ & AES $\uparrow$ & ImgReward $\uparrow$ & MPS $\uparrow$ \\
    \midrule
    SDXL & 27.93 & 5.803 & 0.819 & 14.35 & 27.32 & 5.683 & 0.747 & 11.38 & 27.95 & 5.844 & 0.649 & 11.36 \\
    Diff-DPO & 29.76 & 5.851 & 1.037 & 14.70 & 28.74 & 5.789 & 1.032 & 11.84 & 29.59 & 5.908 & 0.924 & 11.68 \\
    MaPO & 29.05 & 5.925 & 0.925 & 14.46 & 28.03 & 5.851 & 0.864 & 11.46 & 28.88 & 6.001 & 0.794 & 11.33 \\
    SmPO$^*$ & 31.40 & 5.950 & 1.081 & 14.89 & 30.03 & 5.890 & \underline{1.100} & 11.96 & 31.03 & 6.000 & 0.984 & 11.73 \\
    SPO$^*$ & \underline{32.32} & \underline{6.015} & \underline{1.103} & \underline{15.36} & \underline{30.54} & \underline{5.940} & 1.058 & \underline{12.16} & \underline{31.69} & \underline{6.068} & \underline{1.023} & \underline{12.13} \\
    \textbf{CRAFT(Ours)} & \textbf{32.67} & \textbf{6.031} & \textbf{1.312} & \textbf{15.62} & \textbf{31.10} & \textbf{5.976} & \textbf{1.252} & \textbf{12.41} & \textbf{32.18} & \textbf{6.080} & \textbf{1.308} & \textbf{12.53} \\
    \bottomrule
    \end{tabular}
\end{table*}

\begin{figure}[t]
    \centering
    \includegraphics[width=\columnwidth]{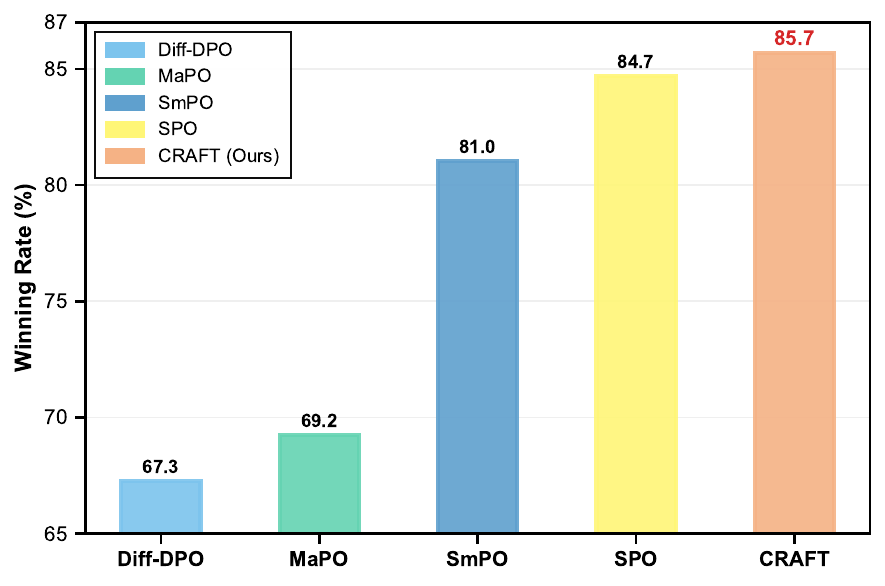}
    \caption{\textbf{Winning Rate Comparison on Parti-Prompt.} CRAFT outperforms all baseline methods in average winning rate against the base SDXL model.}
    \label{fig:win_rate}
\end{figure}

\begin{table*}[t]
\centering
\caption{\textbf{Quantitative Results} of our \textbf{CRAFT} and other baselines for SD1.5. We report the average reward scores of images generated on HPDv2, Parti-Prompt, and Pick-a-Pic datasets. The highest value is shown in \textbf{bold}, the second highest is \underline{underlined}, and models marked with $^*$ are reproduced strictly following the official code and datasets.}
\label{SD1.5_main}
\small
\setlength{\tabcolsep}{2pt}
    \begin{tabular}{c|cccc|cccc|cccc}
    \toprule
    \multirow{2}{*}{\textbf{Model}} & \multicolumn{4}{c|}{\textbf{HPDv2}} & \multicolumn{4}{c|}{\textbf{Parti-Prompt}} & \multicolumn{4}{c}{\textbf{Pick-a-Pic}} \\
    \cmidrule(lr){2-5} \cmidrule(lr){6-9} \cmidrule(lr){10-13}
    & HPS v2.1 $\uparrow$ & AES $\uparrow$ & ImgReward $\uparrow$ & MPS $\uparrow$ & HPS v2.1 $\uparrow$ & AES $\uparrow$ & ImgReward $\uparrow$ & MPS $\uparrow$ & HPS v2.1 $\uparrow$ & AES $\uparrow$ & ImgReward $\uparrow$ & MPS $\uparrow$ \\
    \midrule
    SD1.5 & 24.57 & 5.500 & 0.234 & 12.24 & 25.27 & 5.470 & 0.246 & 9.79 & 24.83 & 5.488 & 0.107 & 9.12 \\
    Diff-DPO & 25.81 & 5.602 & 0.428 & 12.77 & 26.07 & 5.540 & 0.404 & 10.18 & 25.88 & 5.558 & 0.269 & 9.56 \\
    Diff-KTO & \underline{28.56} & 5.681 & \underline{0.740} & \underline{13.12} & \underline{28.10} & 5.667 & \underline{0.641} & 10.28 & \underline{28.10} & 5.662 & \underline{0.624} & 9.55 \\
    SmPO$^*$ & 26.41 & 5.497 & 0.432 & 12.65 & 26.51 & 5.492 & 0.425 & 10.03 & 26.68 & 5.534 & 0.320 & 9.55 \\
    SPO$^*$ & 28.17 & \textbf{5.758} & 0.578 & 13.01 & 27.78 & \textbf{5.718} & 0.531 & \underline{10.42} & 27.78 & \textbf{5.794} & 0.399 & \underline{9.80} \\
    \textbf{CRAFT(Ours)} & \textbf{29.17} & \underline{5.699} & \textbf{0.753} & \textbf{13.36} & \textbf{28.30} & \underline{5.685} & \textbf{0.648} & \textbf{10.48} & \textbf{28.44} & \underline{5.701} & \textbf{0.711} & \textbf{9.90} \\
    \bottomrule
    \end{tabular}
\end{table*}

\paragraph{Evaluation.}
We compare CRAFT with several great fine-tuning approaches, including Diffusion-DPO~\cite{rafailov2023direct}, Diffusion-KTO (for SD1.5)~\cite{li2024aligning}, MaPO (for SDXL)~\cite{hong2024margin}, SmPO~\cite{lu2025smoothed}, and SPO~\cite{liang2025aesthetic}. For a comprehensive evaluation, we adopt three widely used human preference benchmarks: HPDv2 (3,200 prompts)~\cite{wu2023human}, Parti-Prompts (1,632 prompts)~\cite{yu2022scaling}, and the Pick-a-Pic validation set (500 prompts)~\cite{kirstain2023pick}. To further evaluate the general generative capability beyond preference alignment, we test on the Geneval dataset~\cite{ghosh2023geneval}, which covers diverse, open-domain prompts to measure image quality, semantic consistency, and visual diversity.
We evaluate generated images along three complementary dimensions: (1) human preference alignment, (2) generation quality and (3) aesthetic appeal. For human preference alignment, we adopt HPS v2.1~\cite{wu2023human} and employ ImageReward~\cite{xu2024imagereward} to evaluate generation quality. Finally, we evaluate aesthetic appeal using the AES score~\cite{schuhmann2022laion}, which captures visual attractiveness and composition independent of semantic accuracy.
As PickScore~\cite{kirstain2023pick} is trained on the Pick-a-Pic dataset, and most existing alignment methods are optimized on the Pick-a-Pic. To avoid evaluation bias and ensure fair comparison, we exclude PickScore from our evaluation metrics. Furthermore, to demonstrate the generalization of our proposed composite reward beyond the training distribution, we introduce a newly developed metric: MPS~\cite{zhang2024learning}. MPS is a multi-dimensional human preference model that jointly assesses aesthetics, detail quality, semantic alignment, and overall preference, providing a more comprehensive and human-aligned evaluation of text-to-image generation performance. All methods are evaluated under identical generation settings, and we report the average scores across different reward models for each benchmark.

\paragraph{Implementation Details.} For data curation, we first refine an initial set of 10,000 prompts from the HPDv2 training set using Qwen-plus~\cite{bai2023qwen}. For each refined prompt, we generate a group of N=4 images using random seeds. To filter this candidate pool, we apply the $\mathcal{I}_{hpa}$ strategy. The selection and subsequent weighting are guided by a composite reward function $r_{\mathrm{total}}$, which we define as a weighted sum of HPSv2.1, PickScore, and AES with weights of 0.4, 0.4, and 0.2, respectively. The final training dataset is constructed by selecting the highest scoring (text, image) pairs according to this composite reward. We employ AdamW~\cite{loshchilov2017decoupled} as the optimizer and perform full-parameter fine-tuning on the UNet part of the SD1.5 and SDXL models. We train SD1.5 for 120 steps and SDXL for 200 steps with an effective batch size of 128 and a learning rate of 5e-5. All experiments are conducted on 8 H100 GPUs.

\subsection{Main Results}

\paragraph{Qualitative Results.}
We present a qualitative comparison of CRAFT with baseline methods on SDXL in Figure~\ref{fig:qualitative_results}. The results clearly demonstrate CRAFT's superior ability to follow complex prompts. Our method achieves significant improvements across multiple dimensions including text rendering, stylistic interpretation, compositional reasoning, and object fidelity.
For instance, in the first column, CRAFT is the only method that accurately renders the text ``CRAFT" with high aesthetic quality, whereas baselines struggle with spelling and coherence. In the second column, when tasked with a stylistic prompt (``Unreal Engine"), CRAFT delivers a more artistically compelling image that captures the requested aesthetic. Furthermore, CRAFT excels at compositional reasoning. It is the only model to correctly interpret the spatial relationship ``in front of" in the fourth column, and accurately adheres to both color attributes (``white bottle" and ``blue sheep" in column 5) and object counting ( ``four broccolis" in column 6).

\paragraph{Computational Costs.}
As illustrated in Table~\ref{tab:computation_cost_large} and Figure~\ref{fig:efficiency_comparison}, our CRAFT framework demonstrates a dramatic improvement in computational efficiency. While most baselines rely on vast datasets (up to ~928k samples), CRAFT achieves superior alignment using a mere 100 samples. For the fine-tuning stage, CRAFT requires only $\sim$4.0 GPU hours on SDXL and $\sim$1.9 GPU hours on SD1.5. The ``Cost Ratio" column in Table~\ref{tab:computation_cost_large} quantifies this advantage precisely. On SDXL, CRAFT is 15.8× faster than SPO and up to 159.5× faster than Diff-DPO. The efficiency gains are even more pronounced on SD1.5, where CRAFT outperforms SPO by 11.6× and Diff-KTO by a staggering 218.7×. These results confirm that CRAFT not only eliminates the dependency on large-scale preference data but also enables rapid and effective diffusion model alignment at a fraction of the computational cost of existing methods. We attribute this significant improvement in efficiency to the fundamental superiority of our CRAFT framework design.

\paragraph{Quantitative Results.}
Table~\ref{sdxl_main} and Table~\ref{SD1.5_main} summarize the quantitative results of our proposed CRAFT and other baseline methods on SDXL and SD1.5. Each model is evaluated on three test sets: HPDv2, Parti-Prompt, and Pick-a-Pic, using HPSv2.1, AES, ImageReward and MPS. Overall, CRAFT consistently surpasses all baselines across nearly all evaluation metrics and datasets, demonstrating the effectiveness of our composite reward–based fine-tuning strategy. A crucial indicator of our method's generalization capability lies in its performance on metrics Image Reward (IR) and MPS, which are unseen during training. CRAFT demonstrates leading performance on these two metrics, providing compelling evidence that our alignment strategy generalizes effectively rather than merely overfitting to the training rewards. This is particularly evident on the HPDv2 benchmark with the SDXL model, where CRAFT not only attains the highest HPSv2.1 score (32.67) but also achieves the top IR score (1.312) and MPS score (15.62), outperforming the next-best method, SPO. This robust generalization is echoed in the SD1.5 model's results, which leads with an HPSv2.1 of 29.17 and a superior MPS of 13.36. These consistent quantitative advantages are visualized in Figure~\ref{fig:win_rate}, which illustrates CRAFT's dominant average winning rate of over 85.7\% on the Parti-Prompt dataset, underscoring its superior alignment with human preferences.
To further assess compositional reasoning, Table~\ref{geneval_SD1.5} reports the results on the GenEval benchmark. These tests evaluate complex capabilities such as object counting, color and position binding, and attribute association. CRAFT achieves the best or second-best performance across most evaluation categories on SD1.5. This performance on compositional tasks further validates the robust generalization of CRAFT's alignment capabilities beyond standard preference optimization metrics.


\setlength{\tabcolsep}{0.8mm} 
\begin{table}[t]
\centering
\footnotesize 
\caption{\textbf{GenEval results} of our \textbf{CRAFT} and baseline methods for SD1.5. The highest value is shown in bold, the second highest is underlined, and models marked with $^*$ are reproduced strictly following the official code and datasets.}
\label{geneval_SD1.5}
\begin{tabular}{l|ccccccc}
\toprule
           & Single & Two     &         &        &          & Attribute & \\
    Method & Object & Object & Counting & Colors & Position & Binding   & Overall \\
\hline
    SD1.5      & 95.62 & 39.14 & \underline{36.56} & 74.47 & 3.50 & 5.25 & 42.42 \\
    Diff-DPO  & \textbf{98.12} & 38.38 & 35.00 & \underline{77.39} & 4.75 & 6.00 & 43.28 \\
    SmPO$^*$  & 96.25 & 41.92 & 32.19 & 75.53 & 4.00 & \underline{7.25} & 42.86 \\
    SPO$^*$   & 95.94 & \textbf{49.75} & 33.44 & 72.61 & \underline{6.25} & 6.25 & \underline{44.04} \\
    \textbf{CRAFT}  & \underline{96.88} & \underline{47.22} & \textbf{38.12} & \textbf{77.66} & \textbf{6.75} & \textbf{8.25} & \textbf{45.82} \\
\bottomrule
\end{tabular}
\end{table}


\subsection{Ablation and Analysis}

\begin{table}[htbp]
\centering
\caption{\textbf{Data selection strategy comparison.} This table validates our data filtering approach, showing that a small but high-quality dataset is superior. The Top-100 strategy consistently achieves the best performance across all metrics, surpassing strategies that use significantly more, but lower-quality data.}
\label{tab:select_strategy_results}
\setlength{\tabcolsep}{7pt}
\begin{tabular}{l|cc|ccc}
\toprule
\multirow{2}{*}{\textbf{Strategy}} & \multicolumn{2}{c|}{\textbf{Setup}} & \multicolumn{3}{c}{\textbf{Scores}} \\
\cmidrule(lr){2-3} \cmidrule(l){4-6}
& \textbf{Data} & \textbf{Steps} & \textbf{HPS}$\uparrow$ & \textbf{AES}$\uparrow$ & \textbf{IR}$\uparrow$ \\
\midrule
All  & 2,312 & 1,000 & 32.20 & 6.08 & 0.987 \\
Low  & 500   & 500   & 31.86 & 5.94 & 1.040 \\
Rand & 500   & 500   & 33.10 & 6.07 & 1.110 \\
\midrule
\textbf{Top} & \textbf{100} & \textbf{200} & \textbf{33.23} & \textbf{6.13} & \textbf{1.138} \\
\bottomrule
\end{tabular}
\end{table}

\paragraph{Selecting Strategy.}
To evaluate the effectiveness of our composite reward filtering strategy, we conduct an ablation study on data selection. We begin with a set of 2,312 samples, which constitutes the entire dataset curated through our Composite Reward Filtering. We compare four selecting strategies: 1) All: training on all 2,312 samples; 2) Low: training on the 500 lowest-scoring samples; 3) Random: training on 500 randomly selected samples; and 4) Top: training on the 100 highest-scoring samples. Recognizing that larger datasets require more training steps to converge, we scaled the training duration for each strategy to ensure a fair comparison of their peak potential. We report the optimal results achieved for each configuration in Table~\ref{tab:select_strategy_results}. The results clearly demonstrate the superiority of our Top selection strategy. Despite using less than 5\% of the data (100 samples) and lower training cost (200 steps), this strategy achieves the highest scores across all metrics, including an AES of 6.13 and a remarkable IR score of 1.138. Strikingly, this performance surpasses that of the All strategy trained for 1,000 steps. Conversely, training with the entire dataset yields one of the weakest performances. This strongly validates that meticulous selection of high-quality data is critical for achieving optimal model performance.

\paragraph{Composite Reward Models.}
To determine the optimal composition of our reward function, we conducted an ablation study on different reward model combinations, with results detailed in Table~\ref{tab:rm_ablation}. Our analysis reveals that relying on a single reward model is suboptimal. Specifically, using HPS alone yields the weakest performance, while using PickScore alone proves significantly more effective. Interestingly, the combination of HPS and AES fails to surpass the performance of PickScore, underscoring its critical importance for data quality. Based on these insights, our final composite reward function is designed to heavily favor the most effective signals. We assign equal high weight to HPS and PickScore ($\alpha_h = \alpha_p = 0.4$), reflecting their complementary strengths in capturing human preference and text-image alignment, while assigning a smaller weight to AES ($\alpha_a = 0.2$), which primarily measures general aesthetics. The superior performance of this full combination, as shown in the final row, validates this principled weighting strategy and highlights that while the full combination is best, PickScore is the most crucial individual component.

\begin{table}[htbp]
\centering
\caption{\textbf{Reward model combination ablation.} We evaluate the impact of different Composite Reward functions on fine-tuning SDXL. All images are generated with pick-a-pic test sets.}
\label{tab:rm_ablation}
\setlength{\tabcolsep}{9pt} 
\small
\begin{tabular}{ccc | ccc} 
    \toprule
    \multicolumn{3}{c|}{\textbf{Reward Models}} & \multicolumn{3}{c}{\textbf{Performance Metrics}} \\ 
    \cmidrule(r){1-3} \cmidrule(l){4-6} 
    \textbf{HPS} & \textbf{AES} & \textbf{Pic} & \textbf{HPS}$\uparrow$ & \textbf{AES}$\uparrow$ & \textbf{IR}$\uparrow$ \\ 
    \midrule
    \textcolor{green}{\checkmark} & & & 30.25 & 5.94 & 0.791 \\
    & & \textcolor{green}{\checkmark} & 31.71 & 6.04 & 0.973 \\
    \textcolor{green}{\checkmark} & \textcolor{green}{\checkmark} & & 30.93 & 6.00 & 0.834 \\
    \textcolor{green}{\checkmark} & \textcolor{green}{\checkmark} & \textcolor{green}{\checkmark} & \textbf{32.20} & \textbf{6.08} & \textbf{0.987} \\
    \bottomrule
\end{tabular}
\end{table}

\section{Conclusion}
In this work, we address two key challenges in aligning diffusion models: data dependency and high computational cost. We introduce CRAFT, a simple yet effective framework that achieves strong generation alignment using only a handful of samples. CRAFT first constructs a high-quality dataset with a composite reward filtering strategy, and then fine-tunes diffusion models with a weighted supervised objective guided by composite reward. Theoretically, we prove that CRAFT can be viewed as a lower-bound optimization of group-based reinforcement learning, providing a principled link between supervised fine-tuning and reinforcement learning. Empirically, CRAFT fine-tunes SD1.5 and SDXL with only 100 samples, achieving up to 220× faster training while surpassing the alignment quality of existing methods. These results demonstrate that effective diffusion model alignment does not require massive datasets or computationally intensive optimization. 

\paragraph{Acknowledgement.}
This work was supported by the National Natural Science Foundation of China under Grant No. 62506317.

{
    \small
    \bibliographystyle{ieeenat_fullname}
    \bibliography{main}
}




\clearpage
\onecolumn
\setcounter{page}{1}

\begin{center}
   \textbf{\Large CRAFT: Aligning Diffusion Models with Fine-Tuning Is Easier Than You Think} \\ 
   \vspace{0.5em}
   \textbf{\large Supplementary Material}
\end{center}
\vspace{1em}   

\appendix

\section{Detailed Proof of Theorem~\ref{lower bound}}

We provide a fully detailed derivation of the proposed lower bound for clarity.

\subsection*{Step 0: Restate the empirical objective.}

Recall the empirical Monte Carlo objective:
\begin{equation}\label{J hat}
	\hat{J}(\theta)=\mathbb{E}_{\boldsymbol{c}\sim\mathcal{D}_{\boldsymbol{c}},\{\boldsymbol{x}_0^{(i)}\}_{i=1}^G\sim p_{\theta_{\mathrm{old}}}(\cdot|\boldsymbol{c})}\left[\frac{1}{G}\sum_{i=1}^{G}\frac{p_{\theta}(\boldsymbol{x}_0^{(i)}|\boldsymbol{c})}{p_{\theta_{\mathrm{old}}}(\boldsymbol{x}_0^{(i)}|\boldsymbol{c})}\cdot\hat{A}_i\right],
\end{equation}
where
\begin{equation}
	\hat{A}_i=\frac{r(\boldsymbol{c},\boldsymbol{x}_0^{(i)})-\mathrm{mean}\{r(\boldsymbol{c},\boldsymbol{x}_0^{(1)}),\dots,r(\boldsymbol{c},\boldsymbol{x}_0^{(G)})\}}{\mathrm{std}\{r(\boldsymbol{c},\boldsymbol{x}_0^{(1)}),\dots,r(\boldsymbol{c},\boldsymbol{x}_0^{(G)})\}+\epsilon},\quad\frac{1}{G}\sum_{i=1}^{G}\hat{A}_i=0.
\end{equation}

\subsection*{Step 1: Express in terms of log-likelihood difference.}

Define the log-likelihood difference:
\begin{equation}
	\Delta_i(\theta):=\log p_{\theta}(\boldsymbol{x}_0^{(i)}|\boldsymbol{c})-\log p_{\theta_{\mathrm{old}}}(\boldsymbol{x}_0^{(i)}|\boldsymbol{c}).
\end{equation}
Then \eqref{J hat} becomes
\begin{equation}
	\hat{J}(\theta)=\mathbb{E}_{\boldsymbol{c}\sim\mathcal{D}_{\boldsymbol{c}},\{\boldsymbol{x}_0^{(i)}\}_{i=1}^G\sim p_{\theta_{\mathrm{old}}}(\cdot|\boldsymbol{c})}\left[\frac{1}{G}\sum_{i=1}^{G}\exp\left(\Delta_i(\theta)\right)\cdot\hat{A}_i\right].
\end{equation}

\subsection*{Step 2: ELBO form for diffusion models.}

Now according to the evidence lower bound (ELBO) of the log-likelihood in diffusion models \citep{ho2020denoising}, we have
\begin{equation}
	\begin{split}
		\log p_\theta(\boldsymbol{x}_0^{(i)}|\boldsymbol{c})
		&= \log \int p_\theta(\boldsymbol{x}_{0:T}^{(i)}|\boldsymbol{c}) \, \mathrm{d}\boldsymbol{x}_{1:T}^{(i)} \\
		&= \log \int q(\boldsymbol{x}_{1:T}^{(i)}|\boldsymbol{x}_0^{(i)}) \frac{p_\theta(\boldsymbol{x}_{0:T}^{(i)}|\boldsymbol{c})}{q(\boldsymbol{x}_{1:T}^{(i)}|\boldsymbol{x}_0^{(i)})} \mathrm{d}\boldsymbol{x}_{1:T}^{(i)} \\
		&\ge \mathbb{E}_{q(\boldsymbol{x}_{1:T}^{(i)}|\boldsymbol{x}_0^{(i)})} \left[ \log \frac{p_\theta(\boldsymbol{x}_{0:T}^{(i)}|\boldsymbol{c})}{q(\boldsymbol{x}_{1:T}^{(i)}|\boldsymbol{x}_0^{(i)})} \right] \\
		&=-\mathop{\mathbb{E}}_{\substack{t\sim\mathrm{Uniform}(\{1,\dots,T\})\\\boldsymbol{\epsilon}_t^{(i)}\sim\mathcal{N}(0,I)}} \Bigg[
		\underbrace{\frac{1}{2\sigma_t^2} \cdot \frac{1-\bar\alpha_t}{\bar\alpha_t}}_{w(t)}\cdot\left\Vert\boldsymbol{\epsilon}_\theta(\boldsymbol{x}_t^{(i)},t,\boldsymbol{c})-\boldsymbol{\epsilon}_t^{(i)}\right\Vert^2
		\Bigg]+ \text{const}. \\
	\end{split}
\end{equation}
So we can simply denote
\begin{equation}
	\log p_\theta(\boldsymbol{x}_0^{(i)}|\boldsymbol{c})=-\mathop{\mathbb{E}}_{\substack{t\sim\mathrm{Uniform}(\{1,\dots,T\})\\\boldsymbol{\epsilon}_t^{(i)}\sim\mathcal{N}(0,I)}} \Bigg[w(t)\cdot\left\Vert\boldsymbol{\epsilon}_\theta(\boldsymbol{x}_t^{(i)},t,\boldsymbol{c})-\boldsymbol{\epsilon}_t^{(i)}\right\Vert^2
	\Bigg]+C_i,
\end{equation}
where $C_i$ does not depend on \(\theta\). Similarly, for the old parameters:
\begin{equation}
	\log p_{\theta_{\mathrm{old}}}(\boldsymbol{x}_0^{(i)}|\boldsymbol{c})=-\mathop{\mathbb{E}}_{\substack{t\sim\mathrm{Uniform}(\{1,\dots,T\})\\\boldsymbol{\epsilon}_t^{(i)}\sim\mathcal{N}(0,I)}} \Bigg[w(t)\cdot\left\Vert\boldsymbol{\epsilon}_{\theta_{\mathrm{old}}}(\boldsymbol{x}_t^{(i)},t,\boldsymbol{c})-\boldsymbol{\epsilon}_t^{(i)}\right\Vert^2
	\Bigg]+C_i',
\end{equation}
Subtracting gives the exact identity:
\begin{equation}
	\Delta_i(\theta)=-\mathop{\mathbb{E}}_{\substack{t\sim\mathrm{Uniform}(\{1,\dots,T\})\\\boldsymbol{\epsilon}_t^{(i)}\sim\mathcal{N}(0,I)}} \Bigg[w(t)\cdot\left(\left\Vert\boldsymbol{\epsilon}_{\theta_{\mathrm{old}}}(\boldsymbol{x}_t^{(i)},t,\boldsymbol{c})-\boldsymbol{\epsilon}_t^{(i)}\right\Vert^2-\left\Vert\boldsymbol{\epsilon}_{\theta}(\boldsymbol{x}_t^{(i)},t,\boldsymbol{c})-\boldsymbol{\epsilon}_t^{(i)}\right\Vert^2\right)
	\Bigg]+C_i-C_i'.
\end{equation}
Define
\begin{equation}
	M_i^{\theta}:=\mathop{\mathbb{E}}_{\substack{t\sim\mathrm{Uniform}(\{1,\dots,T\})\\\boldsymbol{\epsilon}_t^{(i)}\sim\mathcal{N}(0,I)}}\Bigg[w(t)\cdot\left\Vert\boldsymbol{\epsilon}_\theta(\boldsymbol{x}_t^{(i)},t,\boldsymbol{c})-\boldsymbol{\epsilon}_t^{(i)}\right\Vert^2
	\Bigg],\quad M_i^{\theta_{\mathrm{old}}}:=\mathop{\mathbb{E}}_{\substack{t\sim\mathrm{Uniform}(\{1,\dots,T\})\\\boldsymbol{\epsilon}_t^{(i)}\sim\mathcal{N}(0,I)}}\Bigg[w(t)\cdot\left\Vert\boldsymbol{\epsilon}_{\theta_{\mathrm{old}}}(\boldsymbol{x}_t^{(i)},t,\boldsymbol{c})-\boldsymbol{\epsilon}_t^{(i)}\right\Vert^2
	\Bigg].
\end{equation}
Then $\Delta_i(\theta)=-M_i^{\theta}+M_i^{\theta_{\mathrm{old}}}+C_i-C_i'$.

\subsection*{Step 3: Factor out constants.}

The term $C_i-C_i'$ does not depend on $\theta$, so
\begin{equation}
	\exp\left(\Delta_i(\theta)\right)=\exp\left(C_i-C_i'\right)\cdot\exp\left(-M_i^{\theta}+M_i^{\theta_{\mathrm{old}}}\right),
\end{equation}
where $\exp\left(C_i-C_i'\right)$ can be absorbed into a constant $\tilde{C}$. Hence
\begin{equation}
	\hat{J}(\theta)=\tilde{C}+\mathbb{E}_{\boldsymbol{c}\sim\mathcal{D}_{\boldsymbol{c}},\{\boldsymbol{x}_0^{(i)}\}_{i=1}^G\sim p_{\theta_{\mathrm{old}}}(\cdot|\boldsymbol{c})}\left[\frac{1}{G}\sum_{i=1}^{G}\exp\left(-M_i^{\theta}+M_i^{\theta_{\mathrm{old}}}\right)\cdot\hat{A}_i\right].
\end{equation}

\subsection*{Step 4: Taylor expansion via small learning rate.}

Assume $\theta=\theta_{\mathrm{old}}+\eta g$, with $\eta\rightarrow0$. The noise predictor $\boldsymbol{\epsilon}_{\theta}$ is a smooth function of $\theta$, so we can do a first-order Taylor expansion, i.e., $f(\theta_{\mathrm{old}} + \eta g) = f(\theta_{\mathrm{old}}) + \eta \nabla_\theta f(\theta_{\mathrm{old}}) \cdot g + O(\eta^2)$, we have
\begin{equation}
	\boldsymbol{\epsilon}_{\theta}(\boldsymbol{x}_t^{(i)},t,\boldsymbol{c})=\boldsymbol{\epsilon}_{\theta_{\mathrm{old}}}(\boldsymbol{x}_t^{(i)},t,\boldsymbol{c})+\eta\nabla_\theta\boldsymbol{\epsilon}_{\theta}(\boldsymbol{x}_t^{(i)},t,\boldsymbol{c})|_{\theta_{\mathrm{old}}}\cdot g+O(\eta^2).
\end{equation}
Thus the difference
\begin{equation}
	\boldsymbol{\epsilon}_\theta(\boldsymbol{x}_t^{(i)},t,\boldsymbol{c})-\boldsymbol{\epsilon}_{\theta_{\mathrm{old}}}(\boldsymbol{x}_t^{(i)},t,\boldsymbol{c})=O(\eta).
\end{equation}
Consequently, the squared error inside $M_i^{\theta}$ can be expanded:
\begin{equation}
	\begin{split}
		&\left\Vert\boldsymbol{\epsilon}_\theta(\boldsymbol{x}_t^{(i)},t,\boldsymbol{c})-\boldsymbol{\epsilon}_t^{(i)}\right\Vert^2\\
		=&\left\Vert\boldsymbol{\epsilon}_\theta(\boldsymbol{x}_t^{(i)},t,\boldsymbol{c})-\boldsymbol{\epsilon}_{\theta_{\mathrm{old}}}(\boldsymbol{x}_t^{(i)},t,\boldsymbol{c})+\boldsymbol{\epsilon}_{\theta_{\mathrm{old}}}(\boldsymbol{x}_t^{(i)},t,\boldsymbol{c})-\boldsymbol{\epsilon}_t^{(i)}\right\Vert^2\\
		=&\left\Vert\boldsymbol{\epsilon}_{\theta_{\mathrm{old}}}(\boldsymbol{x}_t^{(i)},t,\boldsymbol{c})-\boldsymbol{\epsilon}_t^{(i)}\right\Vert^2+2\left(\boldsymbol{\epsilon}_{\theta_{\mathrm{old}}}(\boldsymbol{x}_t^{(i)},t,\boldsymbol{c})-\boldsymbol{\epsilon}_t^{(i)}\right)^\top\left(\boldsymbol{\epsilon}_\theta(\boldsymbol{x}_t^{(i)},t,\boldsymbol{c})-\boldsymbol{\epsilon}_{\theta_{\mathrm{old}}}(\boldsymbol{x}_t^{(i)},t,\boldsymbol{c})\right)+O(\eta^2).\\
	\end{split}
\end{equation}
Taking expectation over $t$ and $\boldsymbol{\epsilon}_t^{(i)}$ gives
\begin{equation}
	M_i^{\theta}=M_i^{\theta_{\mathrm{old}}}+O(\eta)\Longrightarrow M_i^{\theta}-M_i^{\theta_{\mathrm{old}}}=O(\eta),
\end{equation}
justifying the linearization:
\begin{equation}
	\exp\left(-M_i^{\theta}+M_i^{\theta_{\mathrm{old}}}\right)=\exp\left(-\left(M_i^{\theta}-M_i^{\theta_{\mathrm{old}}}\right)\right)=1-\left(M_i^{\theta}-M_i^{\theta_{\mathrm{old}}}\right)+O(\eta^2).
\end{equation}

\subsection*{Step 6: Substitute back $\exp\left(-M_i^{\theta}+M_i^{\theta_{\mathrm{old}}}\right)$ in terms of diffusion MSE.}

Finally,
\begin{equation}
	\begin{split}
		\hat{J}(\theta)&=\tilde{C}+\mathbb{E}_{\boldsymbol{c}\sim\mathcal{D}_{\boldsymbol{c}},\{\boldsymbol{x}_0^{(i)}\}_{i=1}^G\sim p_{\theta_{\mathrm{old}}}(\cdot|\boldsymbol{c})}\left[\frac{1}{G}\sum_{i=1}^{G}\exp\left(-M_i^{\theta}+M_i^{\theta_{\mathrm{old}}}\right)\cdot\hat{A}_i\right]\\
		&=\tilde{C}+\mathbb{E}_{\boldsymbol{c}\sim\mathcal{D}_{\boldsymbol{c}},\{\boldsymbol{x}_0^{(i)}\}_{i=1}^G\sim p_{\theta_{\mathrm{old}}}(\cdot|\boldsymbol{c})}\left[\frac{1}{G}\sum_{i=1}^{G}\left(1-\left(M_i^{\theta}-M_i^{\theta_{\mathrm{old}}}\right)+O(\eta^2)\right)\cdot\hat{A}_i\right]\\
		&=\tilde{C}+\underbrace{\mathbb{E}_{\boldsymbol{c}\sim\mathcal{D}_{\boldsymbol{c}},\{\boldsymbol{x}_0^{(i)}\}_{i=1}^G\sim p_{\theta_{\mathrm{old}}}(\cdot|\boldsymbol{c})}\left[\frac{1}{G}\sum_{i=1}^{G}\hat{A}_i\right]}_{=0}+\underbrace{\mathbb{E}_{\boldsymbol{c}\sim\mathcal{D}_{\boldsymbol{c}},\{\boldsymbol{x}_0^{(i)}\}_{i=1}^G\sim p_{\theta_{\mathrm{old}}}(\cdot|\boldsymbol{c})}\left[\frac{1}{G}\sum_{i=1}^{G}M_i^{\theta_{\mathrm{old}}}\cdot\hat{A}_i\right]}_{\mathrm{constant}}\\
		&\quad-\mathbb{E}_{\boldsymbol{c}\sim\mathcal{D}_{\boldsymbol{c}},\{\boldsymbol{x}_0^{(i)}\}_{i=1}^G\sim p_{\theta_{\mathrm{old}}}(\cdot|\boldsymbol{c})}\left[\frac{1}{G}\sum_{i=1}^{G}M_i^{\theta}\cdot\hat{A}_i\right]+O(\eta^2), \\
	\end{split}
\end{equation}
substituting the definition of $M_i^{\theta}$, we have
\begin{equation}
	\hat{J}(\theta)\geq C-\mathop{\mathbb{E}}_{\substack{\boldsymbol{c}\sim\mathcal{D}_{\boldsymbol{c}},\\\{\boldsymbol{x}_0^{(i)}\}_{i=1}^G\sim p_{\theta_{\mathrm{old}}}(\cdot|\boldsymbol{c}),\\t\sim\mathrm{Uniform}(\{1,\dots,T\})\\\boldsymbol{\epsilon}_t^{(i)}\sim\mathcal{N}(0,I)}}\left[\frac{1}{G}\sum_{i=1}^{G}\hat{A}_iw(t)\cdot\left\Vert\boldsymbol{\epsilon}_\theta(\boldsymbol{x}_t^{(i)},t,\boldsymbol{c})-\boldsymbol{\epsilon}_t^{(i)}\right\Vert^2\right].
\end{equation}
This completes the proof.

\section{Supplementary Experimental Results}\label{Supplementary Experimental Results}

\paragraph{Evaluation Bias on PicScore.}

In this work, we exclude PickScore from our primary evaluation metrics to ensure a fair and unbiased comparison. Most state-of-the-art baselines (e.g., Diff-DPO, SPO, SmPO) are explicitly optimized using the Pick-a-Pic dataset. Since the PickScore evaluator is trained on the exact same preference distribution, using it to evaluate these methods introduces significant \textbf{in-domain bias}, where high scores may reflect dataset overfitting rather than generalized generation quality. To avoid this data leakage and demonstrate the robustness of our method, we adopt a comprehensive suite of four independent metrics: \textbf{HPSv2.1}, \textbf{ImageReward}, \textbf{Aesthetic Score (AES)}, and \textbf{MPS}. 
These evaluators are derived from diverse data sources distinct from the baselines' training sets, providing a neutral ground for comparison. By validating our approach across this multi-dimensional framework, we ensure that the reported performance reflects genuine improvements in human alignment and aesthetic quality, fully substantiating the effectiveness of our method without relying on potentially biased indicators.

\paragraph{Geneval Results on SDXL.}
We extend our evaluation to the larger SDXL architecture to verify the scalability and effectiveness of our method. As presented in Table~\ref{tab:geneval_sdxl}, \textbf{CRAFT} achieves the highest `Overall' score of \textbf{57.97}, surpassing the base SDXL model (55.05) as well as recent alignment baselines such as Diff-DPO (57.23) and SmPO (57.86). Notably, CRAFT demonstrates superior capability in ensuring object presence and spatial fidelity. It secures state-of-the-art results in the `Single Object' (\textbf{99.06}) and `Two Object' (\textbf{86.36}) categories. The significant margin in the `Two Object' metric, outperforming the second-best method by nearly 6 points, highlights CRAFT's robustness in complex multi-subject generation scenarios. Furthermore, CRAFT also leads in the `Position' category with a score of \textbf{14.50}, while maintaining highly competitive performance in attribute binding and color consistency. These results confirm that CRAFT effectively enhances the controllability and compositional alignment of large-scale text-to-image models.

\begin{figure}[t]
    \centering
    \includegraphics[width=\columnwidth]{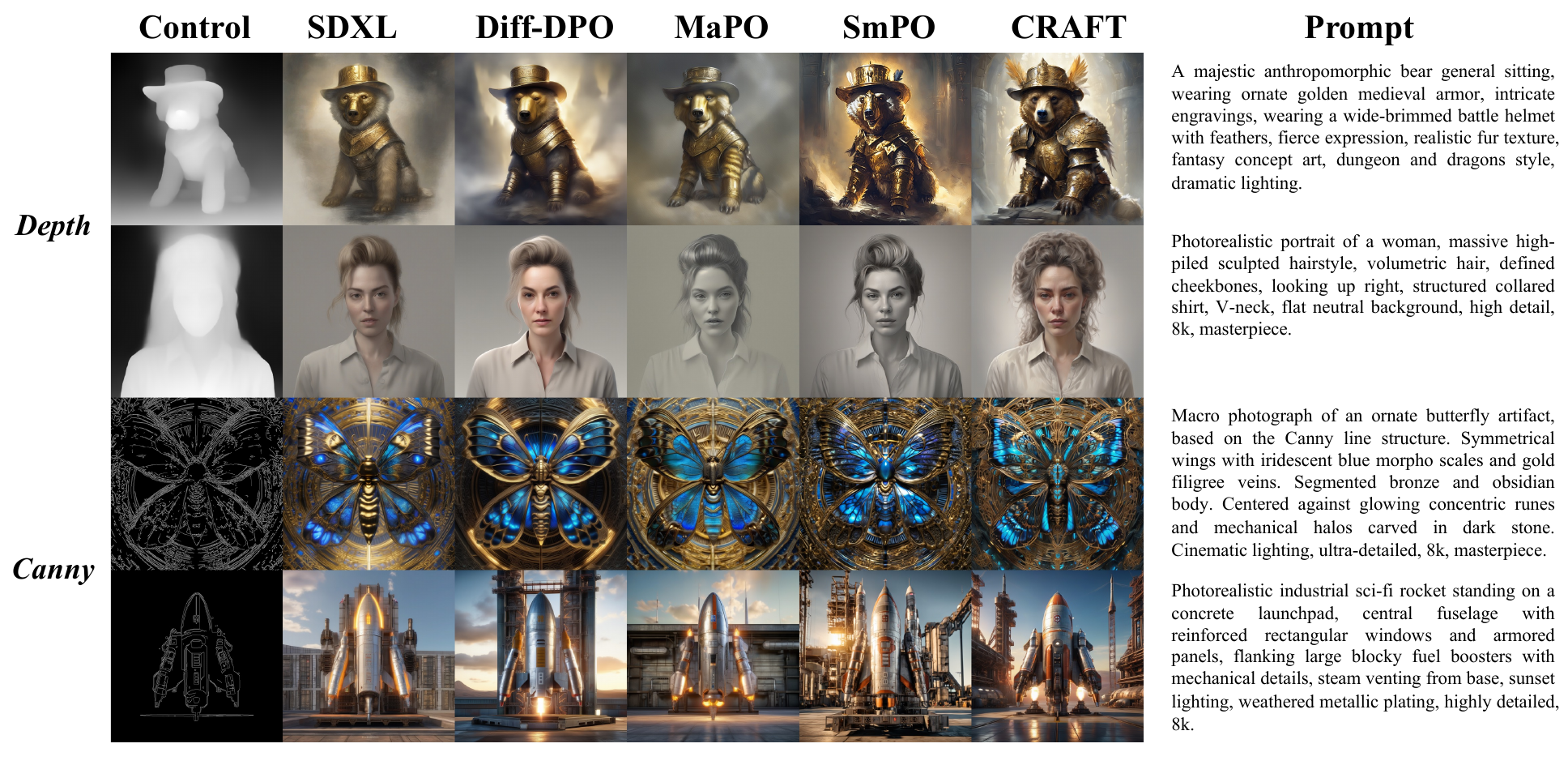}
    \caption{\textbf{ControlNet Visualization: Superior Control Fidelity and Data Efficiency of CRAFT.} 
Qualitative comparison using diverse ControlNet conditions (\textbf{Canny} and \textbf{Depth}) against existing fine-tuning methods (SDXL, Diff-DPO, MaPO, SmPO). 
CRAFT consistently demonstrates strong geometric control fidelity and generates visually superior images across all modalities. 
This high robustness and aesthetic quality are achieved despite utilizing significantly fewer training samples, highlighting CRAFT's superior data efficiency and generalization capability.
}
\label{fig:controlnet_comparison}
\end{figure}

\begin{figure}[t]
    \centering
    \includegraphics[width=\columnwidth]{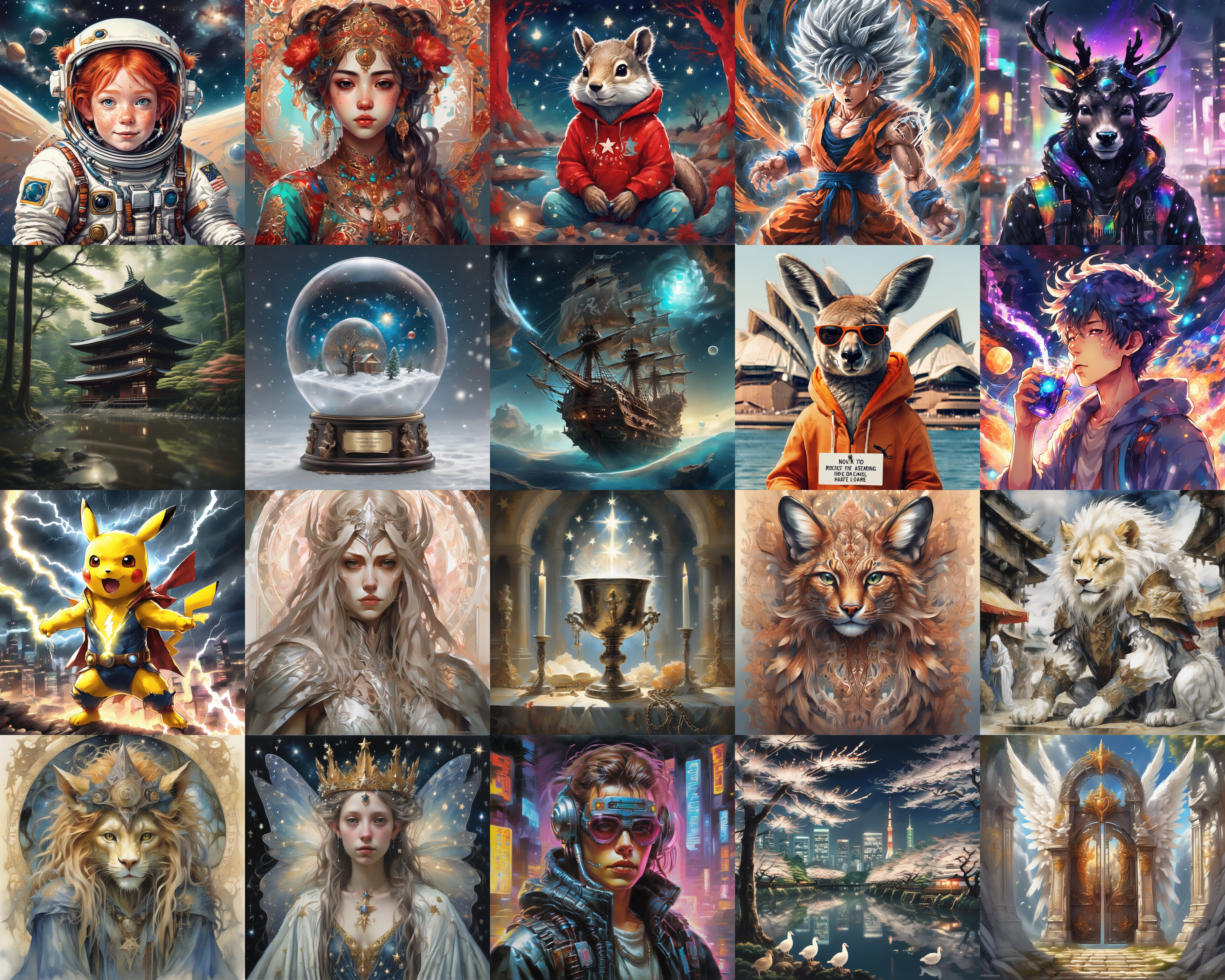}
    \caption{\textbf{Qualitative results of CRAFT-SDXL.} After CRAFT fine-tuning, the model demonstrates the capability to generate visually superior images with exceptional aesthetic quality.}
    \label{fig:stitched_result_1}
\end{figure}

\paragraph{More Visulization.}
Figure~\ref{fig:stitched_result_1} provide additional visually appealing samples generated by CRAFT-SDXL, showcasing both its general aesthetic superiority and its specific structural integrity. Consistent with the main text, the prompts for these images are sourced from the HPDv2, Parti-Prompt, and Pick-a-Pic test sets, affirming performance across diverse domains. Figure~\ref{fig:controlnet_comparison} includes results from ControlNet, demonstrating CRAFT's superior control fidelity when guided by conditioning inputs (such as Canny and Depth maps). These samples confirm the model's robustness: the high aesthetic quality is consistently maintained even when the geometric structure is strictly constrained by external control signals, which is a key indicator of model generalization beyond simple style transfer.

\paragraph{End-to-End Efficiency.} To provide a fairer efficiency comparison, we report the end-to-end cost of our pipeline, including candidate generation, reward-based filtering, and the final fine-tuning stage. As shown in Table~\ref{tab:computation_cost_e2e}, CRAFT remains substantially more efficient than existing baselines even after including preprocessing overhead. This is important because our method does not assume access to large-scale preference pairs; instead, it constructs a compact training set from prompts with a comparatively light self-curation stage.

\begin{table}[t]
\centering
\caption{\textbf{End-to-end cost in H100 hours.} For CRAFT, we report both preprocessing and fine-tuning cost. Even with candidate generation and filtering included, CRAFT remains substantially more efficient than prior methods.}
\label{tab:computation_cost_e2e}
\vspace{-6pt}
\scriptsize
\setlength{\tabcolsep}{4pt}
\resizebox{0.86\linewidth}{!}{%
\begin{tabular}{lcc|lcc}
\toprule
\multicolumn{3}{c|}{\textbf{SDXL}} & \multicolumn{3}{c}{\textbf{SD1.5}} \\
\cmidrule(lr){1-3} \cmidrule(lr){4-6}
\textbf{Method} & \textbf{Hours} & \textbf{Ratio}
& \textbf{Method} & \textbf{Hours} & \textbf{Ratio} \\
\midrule
Diff-DPO  & $\sim$638.1 & 26.5$\times$
& Diff-DPO & $\sim$80.6 & 9.2$\times$ \\
MaPO & $\sim$545.6 & 22.6$\times$
& Diff-KTO & $\sim$415.6 & 47.2$\times$ \\
\textbf{CRAFT} 
& $\sim$(4.0 + 20.1) & \textbf{1.0$\times$}
& \textbf{CRAFT}
& $\sim$(1.9 + 6.9) & \textbf{1.0$\times$} \\
\bottomrule
\end{tabular}}
\vspace{-8pt}
\end{table}

\paragraph{Additional Preference-Optimization Baselines.} We further compare CRAFT against recent preference-optimization baselines. Table~\ref{tab:inpo_metrics} reports score-based comparisons against InPO using its public checkpoint. Table~\ref{tab:dspo_winrate} summarizes the comparison against DSPO~\cite{zhu2025dspo} using the reward win-rate statistics reported in its original paper, since official checkpoints are unavailable. Across all reported datasets and metrics, CRAFT remains consistently stronger.

\begin{table}[t]
\centering
\scriptsize
\setlength{\tabcolsep}{2pt}
\begin{minipage}[t]{0.48\linewidth}
\centering
\caption{\textbf{Comparison with InPO on SDXL.}}
\label{tab:inpo_metrics}
\vspace{-8pt}
\resizebox{\linewidth}{!}{%
\begin{tabular}{l|ccc|ccc|ccc}
\toprule
\multirow{2}{*}{\textbf{Model}}
& \multicolumn{3}{c|}{\textbf{HPDv2}}
& \multicolumn{3}{c|}{\textbf{Parti-Prompt}}
& \multicolumn{3}{c}{\textbf{Pick-a-Pic}} \\
\cmidrule(lr){2-4} \cmidrule(lr){5-7} \cmidrule(lr){8-10}
& HPS & AES & IR
& HPS & AES & IR
& HPS & AES & IR \\
\midrule
InPO
& 30.79 & 5.837 & 1.058
& 29.27 & 5.756 & 1.024
& 30.30 & 5.870 & 0.978 \\
\textbf{CRAFT}
& \textbf{32.67} & \textbf{6.031} & \textbf{1.312}
& \textbf{31.10} & \textbf{5.976} & \textbf{1.252}
& \textbf{32.18} & \textbf{6.080} & \textbf{1.308} \\
\bottomrule
\end{tabular}}
\end{minipage}
\hfill
\begin{minipage}[t]{0.48\linewidth}
\centering
\caption{\textbf{Comparison with DSPO on SDXL.}}
\label{tab:dspo_winrate}
\vspace{-8pt}
\resizebox{\linewidth}{!}{%
\begin{tabular}{l|ccc|ccc|ccc}
\toprule
\multirow{2}{*}{\textbf{Model}}
& \multicolumn{3}{c|}{\textbf{HPDv2}}
& \multicolumn{3}{c|}{\textbf{Parti-Prompt}}
& \multicolumn{3}{c}{\textbf{Pick-a-Pic}} \\
\cmidrule(lr){2-4} \cmidrule(lr){5-7} \cmidrule(lr){8-10}
& HPS & AES & IR
& HPS & AES & IR
& HPS & AES & IR \\
\midrule
DSPO
& 83.47 & 51.41 & 70.09
& 81.80 & 57.84 & 73.47
& 80.00 & 54.20 & 68.60 \\
\textbf{CRAFT}
& \textbf{97.84} & \textbf{73.00} & \textbf{84.38}
& \textbf{94.73} & \textbf{81.50} & \textbf{80.70}
& \textbf{96.40} & \textbf{74.00} & \textbf{86.00} \\
\bottomrule
\end{tabular}}
\end{minipage}
\vspace{-8pt}
\end{table}

\paragraph{Isolating the Effect of Advantage Weighting.} To separate the contribution of data selection from the contribution of the training objective, we train a vanilla SFT baseline on the exact same Top-100 filtered dataset used by CRAFT. Table~\ref{tab:ablation_sft_appendix} shows that advantage weighting yields clear gains over vanilla SFT, especially on HPS and ImageReward, demonstrating that the performance gain does not come solely from data filtering.

\begin{table}[t]
\centering
\scriptsize
\setlength{\tabcolsep}{2pt}
\begin{minipage}[t]{0.47\linewidth}
\centering
\caption{\textbf{Vanilla SFT vs. CRAFT.}}
\label{tab:ablation_sft_appendix}
\vspace{-8pt}
\resizebox{0.75\linewidth}{!}{%
\begin{tabular}{l|ccc}
\toprule
Method & HPS & AES & IR \\
\midrule
Vanilla SFT & 30.91 & 6.040 & 1.110 \\
\textbf{CRAFT} & \textbf{32.18} & \textbf{6.080} & \textbf{1.308} \\
\bottomrule
\end{tabular}}
\end{minipage}
\hfill
\begin{minipage}[t]{0.47\linewidth}
\centering
\caption{\textbf{T2I-CompBench.}}
\label{tab:t2i_compbench_appendix}
\vspace{-8pt}
\resizebox{0.94\linewidth}{!}{%
\begin{tabular}{l|ccc|cc|c}
\toprule
Model
& \multicolumn{3}{c|}{Attr. Bind.}
& \multicolumn{2}{c|}{Obj. Rela.}
& Complex \\
\cmidrule(lr){2-4} \cmidrule(lr){5-6}
& C & S & T
& Sp. & N-Sp.
&  \\
\midrule
SD1.5
& 0.384 & 0.376 & 0.407
& 0.102 & 0.309
& 0.391 \\
\textbf{CRAFT}
& \textbf{0.495} & \textbf{0.446} & \textbf{0.456}
& \textbf{0.139} & \textbf{0.311}
& \textbf{0.405} \\
\bottomrule
\end{tabular}}
\end{minipage}
\vspace{-8pt}
\end{table}

\begin{figure}[t]
    \centering
    \includegraphics[width=0.82\linewidth]{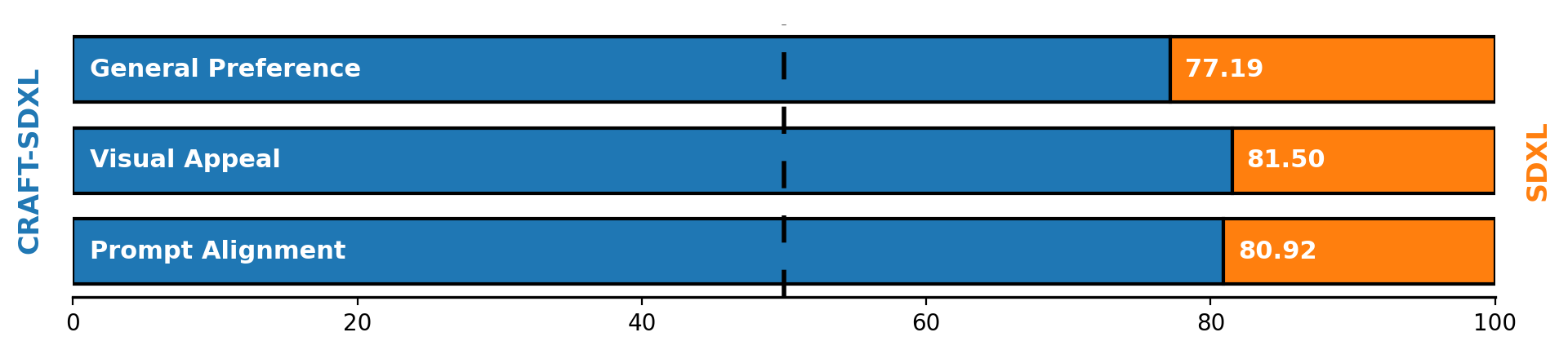}
    \caption{\textbf{User study results.} Human evaluation corroborates the automatic reward-based metrics and favors CRAFT over the compared baselines.}
    \label{fig:user_study_appendix}
    \vspace{-8pt}
\end{figure}

\paragraph{Human Evaluation and Semantic Faithfulness.} Since pretrained reward models would be biased, we additionally report a user study in Figure~\ref{fig:user_study_appendix}. The human preference trend is consistent with our automatic evaluations, providing complementary evidence that the observed gains are not merely artifacts of the reward models. We also report T2I-CompBench in Table~\ref{tab:t2i_compbench_appendix}, which supports that prompt refinement does not degrade compositional or semantic faithfulness.

\paragraph{Prompt Refinement, Reward Balancing, and Dataset Diversity.} We use Qwen-Plus to rewrite original prompt into several more descriptive variants, which are used only for candidate generation. Importantly, all reward computation and filtering are performed with respect to the original prompt, so candidates with semantic drift are naturally disfavored during selection. For composite reward filtering, we first map the three reward signals to comparable scales before applying the final weights $(0.4, 0.4, 0.2)$ for HPS, PickScore, and AES. These coefficients should therefore be interpreted as relative importance weights rather than raw-scale coefficients. The ablation results in Table~\ref{tab:rm_ablation} indicate that the full combination is the most reliable in practice, while Table~\ref{tab:select_strategy_results} shows that a compact Top-100 subset is sufficient for strong performance. In the final selected set, we did not observe exact duplicate prompts, and the strong performance across HPDv2, Parti-Prompt, Pick-a-Pic, GenEval, and T2I-CompBench suggests that the few-shot setting does not bias the model toward a narrow prompt pattern.

\section{Discussion}\label{dis}
\paragraph{Limitations.} Despite the promising results achieved by CRAFT, we acknowledge two primary limitations in our current framework. First, regarding the algorithmic paradigm, CRAFT fundamentally operates as an offline Supervised Fine-Tuning approach. While efficient, it relies on a static dataset constructed prior to training. This formulation inherently limits the model's data utilization efficiency and exploration capability compared to online Reinforcement Learning (RL) strategies (e.g., PPO or the recent GRPO), which can dynamically explore the generation space and potentially reach a higher performance ceiling. Second, regarding the data construction strategy, our current framework still selects training samples at the image level after generation, rather than explicitly modeling the quality of the initial noise itself. Recent studies suggest that high-quality or prompt-aware noise can be selected, optimized, or learned to consistently improve sample quality~\cite{qi2024not, zhou2025golden}. Incorporating such noise-aware criteria into our reward filtering pipeline is a promising yet currently unexplored direction. Third, regarding the application scope, our current experimental validation is exclusively confined to Text-to-Image (T2I) generation. Although the principles of preference-free alignment are theoretically universal, we have not yet addressed the unique challenges present in other modalities, such as the temporal consistency required for Text-to-Video (T2V) or the geometric constraints essential for Text-to-3D generation.

\paragraph{Future Directions.} Building upon these observations, our future work will focus on three strategic expansions. To transcend the offline limitations, we aim to evolve CRAFT into an \textit{online} learning framework. By implementing an iterative "generate-train" loop, we can continuously update the training data with the model's own evolving distributions. This on-policy approach will ensure the model receives sustained and increasingly precise gradient signals throughout the training process, bridging the gap between SFT and RL. We also plan to enrich our filtering stage with adaptive noise selection criteria, where the filtering form and quality standards can vary across prompts, models, or downstream objectives, potentially benefiting from learned noise priors and reflection-based sampling signals~\cite{bai2025weak, bai2025weak, qi2024not, zhou2025golden}. Finally, to broaden the application scope, we plan to adapt the CRAFT algorithm to a wider array of generative tasks. Specifically, we will investigate how our composite reward filtering and fine-tuning can be integrated into T2V and T23D pipelines, exploring whether the robust alignment capabilities demonstrated in 2D images can effectively generalize to complex temporal and spatial dimensions. In particular, recent video inference and distillation techniques, such as weak-to-strong video distillation and mixed image-video samplers~\cite{shao2025magicdistillation, shao2024iv}, suggest that alignment-time fine-tuning and video-specific inference improvements may be fruitfully combined.

\paragraph{Broader Opportunities.} Our work is also connected to several broader directions in the diffusion literature. A recent survey summarizes the fundamentals, challenges, and future directions of diffusion alignment~\cite{liu2026alignment}. From the data perspective, diffusion dataset condensation and dataset pruning suggest that compact yet informative training subsets can substantially reduce training cost~\cite{huang2025diffusion, yang2022dataset}, which is consistent with our motivation of aligning models from extremely limited filtered data. From the evaluation perspective, recent work shows that T2I comparisons can be distorted by guidance-scale bias~\cite{xie2026guidance}, reinforcing our use of multiple complementary metrics rather than relying on a single score. Meanwhile, CRAFT focuses on improving model alignment during fine-tuning, but it is naturally compatible with a range of orthogonal inference-time advances. Recent methods improve generation quality or efficiency through collect-reflect-refine pipelines, sparse attention approximations, or carefully designed inference heuristics for alternative generative backbones~\cite{shao2025core, li2026pisa, shao2024bag}. Beyond alignment and synthesis, diffusion models have also shown promise for zero-shot retrieval and classification~\cite{10888226, qi2024simple}. Understanding how these data-centric, inference-centric, and downstream semantic capabilities interact with preference-aligned diffusion models is another valuable direction for future research.

\begin{table}[t!] 
    \centering
    \caption{\textbf{GenEval results} of our \textbf{CRAFT} and baseline methods for SDXL. The highest value is shown in bold, the second highest is underlined, and models marked with $^*$ are reproduced strictly following the official code and datasets.}
    \label{tab:geneval_sdxl}

    \setlength{\tabcolsep}{4pt} 
    
    \resizebox{\linewidth}{!}{
        \begin{tabular}{lccccccc}
            \toprule
            & Single & Two & & & & Attribute & \\
            Method & Object & Object & Counting & Colors & Position & Binding & Overall \\
            \midrule
            SDXL          & 97.50 & 70.96 & 42.81 & \textbf{88.30} & 11.00 & 21.00 & 55.05 \\
            Diff-DPO      & \underline{98.75} & \underline{80.56} & \textbf{45.62} & 86.70 & 11.00 & 20.75 & 57.23 \\
            SmPO$^*$      & \underline{98.75} & 79.55 & \underline{44.69} & 86.70 & 10.50 & \textbf{27.00} & \underline{57.86} \\
            SPO$^*$       & 97.81 & 80.05 & 38.44 & 84.04 & \underline{11.75} & 20.50 & 55.43 \\
            \textbf{CRAFT} (Ours) & \textbf{99.06} & \textbf{86.36} & 36.88 & \underline{87.23} & \textbf{14.50} & \underline{23.75} & \textbf{57.97} \\
            \bottomrule
        \end{tabular}
    }
\end{table}

\end{document}